\documentclass[runningheads]{llncs}
\usepackage{subcaption}
\usepackage{multirow}
\usepackage{graphicx}

\usepackage{times}
\usepackage{latexsym}
\usepackage{graphicx}

\usepackage{tabularx}
\usepackage{booktabs}
\usepackage{multirow}
\usepackage{graphicx}
\usepackage{amsmath}
\usepackage{comment}
\usepackage{pgfplots,wrapfig}
\usepackage{color}
\usepackage{xcolor,colortbl}
\usepackage{enumitem}
\usepackage{ amssymb }
\usepackage{array}

\newcommand{\ra}[1]{\renewcommand{\arraystretch}{#1}}
\newcolumntype{P}[1]{>{\centering\arraybackslash}p{#1}}
\newcolumntype{L}[1]{>{\raggedright\arraybackslash}p{#1}}
\pgfplotsset{compat=1.12}

\begin{document}
\title{CruzAffect at AffCon 2019 Shared Task: \\A feature-rich approach to characterize happiness}
%
\titlerunning{CruzAffect: A feature-rich approach to characterize happiness}
%
\author{Jiaqi Wu \and
Ryan Compton \and Geetanjali Rakshit \and Marilyn Walker \and Pranav Anand \and Steve Whittaker }
\authorrunning{J. Wu et al.}
%
\institute{UC Santa Cruz, 1156 High Street, Santa Cruz 95064, California\\
\email{\{jwu64,rcompton,grakshit,mawalker,panand,swhittak\}@ucsc.edu}}
\maketitle              
\begin{abstract}
We present our system, \textit{CruzAffect}, for the CL-Aff Shared Task
2019. Cruz-Affect consists of several types of robust and efficient
models for affective classification tasks. We utilize both traditional
classifiers, such as XGBoosted Forest, as well as a deep learning
Convolutional Neural Networks (CNN) classifier. We explore rich
feature sets such as syntactic features, emotional features, and
profile features, and utilize several sentiment lexicons, to discover
essential indicators of social involvement and control that a subject
might exercise in their happy moments, as described in textual
snippets from the HappyDB database. The data comes with a labeled set
(10K), and a larger unlabeled set (70K). We therefore use supervised
methods on the 10K dataset, and a bootstrapped semi-supervised
approach for the 70K. We evaluate these models for binary
classification of \textit{agency} and \textit{social} labels (Task 1),
as well as multi-class prediction for \textit{concepts} labels (Task
2). We obtain promising results on the held-out data, suggesting that
the proposed feature sets effectively represent the data for affective
classification tasks. We also build \textit{concepts} models that
discover general themes recurring in happy moments. Our results
indicate that generic characteristics are shared between the classes
of \textit{agency}, \textit{social} and \textit{concepts}, suggesting
it should be possible to build general models for affective
classification tasks.

\end{abstract}
\section{Introduction}
The overall goal of the CL-Aff Shared Task \cite{overview_claff} is to
understand what makes people happy, and the factors contributing
towards such happy moments. Related work has centered around
understanding and building lexicons that focus on emotional
expressions \cite{mohammad2010emotions,tausczik2010psychological},
while Reed {\it et al.}  \cite{Reed17} learn lexico-functional
  linguistic patterns as reliable predictors for first-person affect,
  and constructed a First-Person Sentiment Corpus of positive and
  negative first-person sentences from blog journal entries. Wu {\it
    et al.} \cite{Wu17} propose a synthetic categorization of
  different sources for well-being and happiness targeting the private
  micro-blogs in Echo, where users rate their daily events from 1 to
  9. These work aim to identify specific compositional semantics that
  characterize the sentiment of events, and attempt to model happiness
  at a higher level of generalization, however finding generic
  characteristics for modeling well-being remains challenging. In this
  paper, we aim to find  generic characteristics shared
  between different affective classification tasks. Our
  approach is to compare state-of-the-art methods for linguistic
  modeling to prior lexicons\rq{} predictive power.  While this body
  of work is broader in scope than the goals we are trying to address,
  they do include annotated sets of words associated with happiness as
  well as additional categories of psychological significance.

The aim of this work is to address the two tasks that are part of the CL-Aff Shared Task. The data provided for this task comes from the HappyDB dataset \cite{happydb}. Task 1 focuses on binary prediction of two different labels, \textit{social} and \textit{agency}. The intention is to understand the context surrounding happy moments and potentially find factors associated with these two labels. Task 2 is fairly open-ended, leaving it to the participant's imagination to model happiness and derive insights from their models. Here, we predict the \textit{concepts} label using multi-class classification. We explore various approaches to determine which models work best to characterize contextual aspects of happy moments.
Though the predictions of \textit{agency} and \textit{social} sound
simpler than \textit{concepts}, we expect that the best models for
\textit{agency} and \textit{social} prediction could generate
similarly optimal performance for \textit{concepts}, assuming that the
classes of \textit{social}, \textit{agency}, and \textit{concepts}
share common characteristics. To validate our assumptions, we build
different models for general affective classification tasks and then
try to gain a deeper understanding of the characteristics of happy
moments by interpreting such models with the Riloff's Autoslog
linguistic-pattern learner \cite{Riloff96,Wu17}.

\section{Agency and Social Classification}
This work utilizes a bootstrapping approach to conduct semi-supervised
learning experiments. This involves a three-step procedure: (1)
train a model on the labeled data; (2)  use the trained model to
make predictions on the unlabeled data; and (3) train a new model
using the combination of the labeled data and the predictions on the
unlabeled data. Training each model involves a 10-fold
cross-validation to evaluate the performance, while guaranteeing that the test set for each fold consists of gold-standard hand-labelled instances.

\subsection{Feature Extraction}
We explore different features to find those most informative for the
prediction task. We aim to understand how syntactic features
and emotional features compare to word embeddings, and whether
the profile features improve the prediction results. 

\textbf{Syntactic Features:} Our syntactic features are limited to
\emph{Part of Speech (POS)} tagging, by applying a POS tagger to count the
relative frequencies of syntactic nouns, verbs, adjectives and
adverbs, use of questions as well as tense and aspect information
\cite{toutanova2003feature}. There are 36 POS features.

\textbf{Emotional Features:} We use 4 different types
of emotional features. \emph{LIWC v2007} \cite{tausczik2010psychological} is a lexicon providing frequency counts of words indexing important psychological constructs, as well as relevant topics (Leisure, Work). 
The \emph{Emotion Lexicon (EmoLex)} \cite{mohammad2010emotions} contains 14,182 words classified into 10 emotional categories: Anger, Anticipation, Disgust, Fear, Joy, Negative, Positive, Sadness, Surprise, and Trust. The \emph{Subjectivity Lexicon} is part of OpinionFinder \cite{wilson2005opinionfinder}. It consists of 8222 stemmed and unstemmed words, annotated by a group of trained annotators as either strongly or weakly subjective.
Our last feature is our own regression model from prior work on predicting the \emph{level of factual and emotion language}. Details about this model can be found in \cite{compton2017facts}. There are 94 features in total.

\textbf{Word Embedding:} We utilize \emph{GloVe}
\cite{pennington2014glove} 100 dimension word vectors for word
representation. GloVe is expected to encode distributional
aspects of meaning. 

\textbf{Profile Features:} The corpus include demographic features
collected via a survey: \emph{age}, \emph{country}, \emph{gender},
\emph{married}, \emph{parenthood}, \emph{reflection}, and
\emph{duration}. To reduce sparsity, we
convert the \emph{country} feature into \emph{language} feature,
assuming that the people who speak the same language might share
similar culture, and thus similar happy moments. After this
conversion, we have 48 different languages from the 70 countries, the
largest group is English ($80.3\%$), then Hindi ($15.9\%$), 
corresponding to  79.8\% examples from USA and 15.87\% examples
from IND. We also bin  age into groups, assuming that different age
groups would have different general happy moments. The age groups,
illustrated in Table~\ref{age_table}, include kid ($age <
10$), teenager ($age < 18$), youth ($age < 24$), young adult ($age <
40$), middle age ($age < 65$) and elderly ($age >= 65$). There are 70
features after the feature preprocessing. We aim to test
whether the features extracted from text are sufficient
for affective sentiment analysis, and whether the profile features
improve performance.

\begin{table}
\centering
\caption{The distribution of the age groups, and the probability of P(agency=yes) and P(social=yes) for each age group. The overall probability of P(agency=yes) is 0.74 and P(social=yes) is 0.53. The middle-age group is less likely to identify their happy moment with \textit{agency} but more likely to identify the moment with \textit{social}, while kids are more likely to identify their happy moment with both \textit{agency} and \textit{social}.}\label{age_table}
\begin{small}
\begin{tabular}{|l| c| c| c|} 
\hline
\bfseries age & \bfseries frequency & \bfseries P(agency=yes) & \bfseries P(social=yes)  \\ 
\hline
elderly ($age >= 65$)  &     136      &    0.71        &   0.55          \\ 
\hline
middle-age ($age < 65$) &    1756     &     0.68       &  0.57           \\ 
\hline
adult ($age < 40$)     &    7242     &     0.74       & 0.53            \\ 
\hline
young ($age < 24$)     &     1401      &      0.79      & 0.50            \\ 
\hline
teenager ($age < 18$) &      0     &  0     &   0          \\ 
\hline
kid ($age < 10$) &      14     &     0.86       &    0.79         \\ 
\hline
NaN        &       11    &    0.55        &   0.82          \\
\hline
\end{tabular}
\end{small}
\end{table}

\subsection{Classification Models}
In this section, we compare the performance of traditional machine learning methods, such as Logistic Regression and XGBoosted Random Forest \cite{chen2016xgboost} and a Convolutional Neural Network (CNN) model \cite{ZhangW15b}.

\subsubsection{Supervised Learning.}

\begin{figure}[ht]
\centering
\includegraphics[width=4.8in]{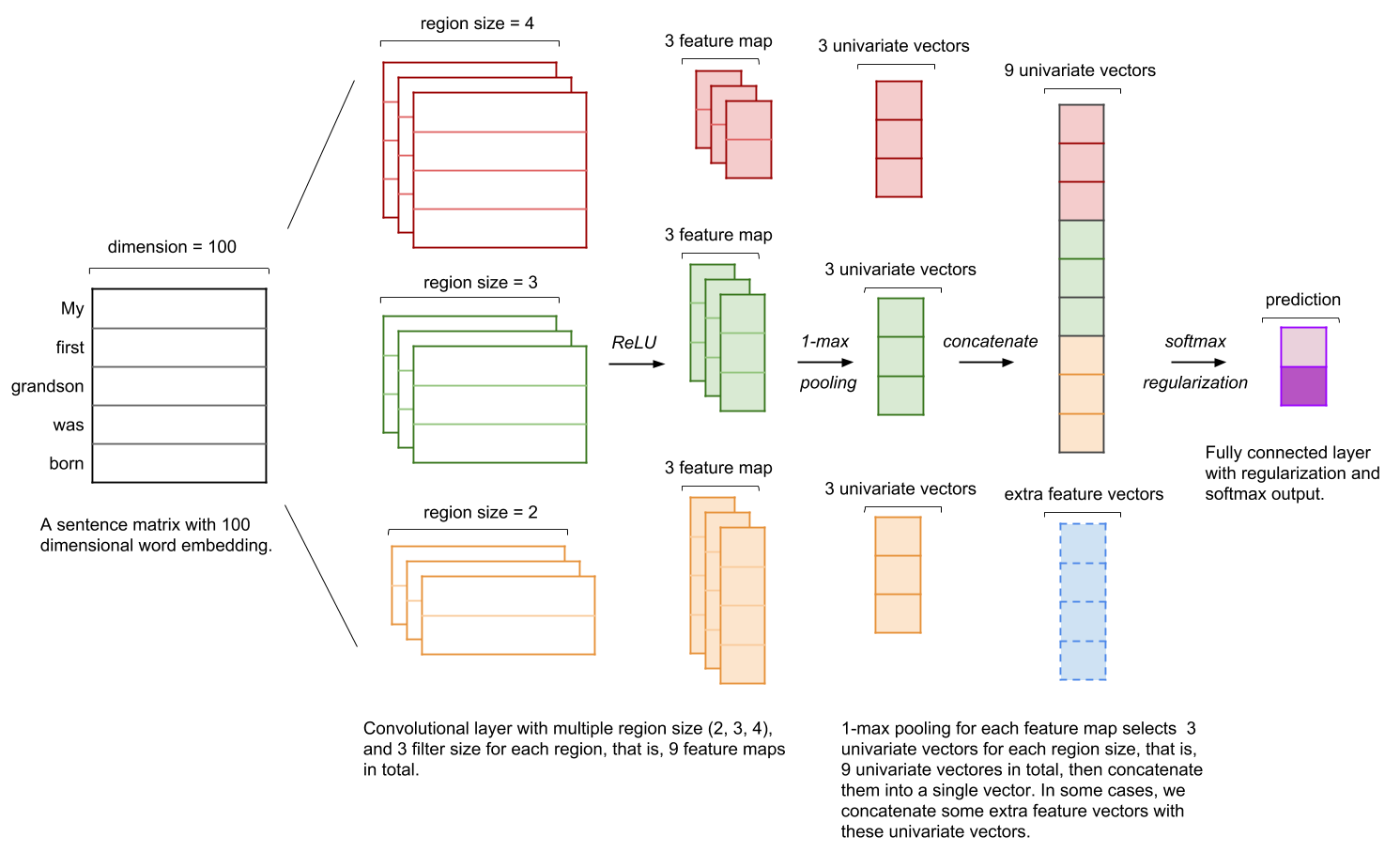}
\caption{A Diagram for the CNN model with region size (2, 3, 4) and filter size 3 for a single sentence. This architecture is cited from \cite{ZhangW15b}. We have tuned some hyperparameters for our tasks. In some cases, we concatenate extra feature vectors, such as the syntactic features and the emotional features that are extracted from a sentence to the univariate vectors, then forward it to the softmax layer for the output.  \label{cnn-diagram-fig}}
\end{figure}

For modeling the profile features, we apply logistic regression with a
liblinear solver and balanced class weights. For modeling the
syntactic features and emotional features, we use  XGBoosted
Random Forest with out-of-the-box values for all
parameters except for the following: 250 number of estimators,
a learning rate of 0.05, and a maximum tree depth of  6. For
the CNN model with word embedding, we explore its performance with
different parameter settings. The best hyperparameters of the CNN
model include filter size 3, multiple region size (2, 3, 4) and max
pooling size 1, or filter size 4, multiple region size (2, 3, 4, 5)
and max pooling size 1. The region size implies a windows size for
N-grams. After getting the best hyperparameters, we train the model
with word embeddings, and word embeddings concatenated with syntactic
and emotional features, to test whether syntactic
and emotional features improve 
performance. Figure~\ref{cnn-diagram-fig} illustrates
the CNN model with region size (2, 3, 4).

\begin{table}
\centering
\caption{10-fold cross-validation supervised learning for \textit{agency} and \textit{social} prediction.}
\label{sup_agency_social}
\begin{small}
\begin{tabular}{| *{10}{c|} }
\hline
\bfseries \multirow{2}{*}{Model} & \bfseries \multirow{2}{*}{Features} & \multicolumn{4}{c|}{\textbf{Agency}}  & \multicolumn{4}{c|}{\textbf{Social}}
\\
\cline{3-10}
 & & \textbf{P} & \textbf{R} & \textbf{F1} & \textbf{Acc} & \textbf{P} & \textbf{R} & \textbf{F1} & \textbf{Acc}
\\
\hline
Logistic Regression  & Profile & 0.55 & 0.57 & 0.53 & 0.57 & 0.56 & 0.56 & 0.55 & 0.56
\\\hline
XGBoosted Forest & Syn. \& Emo.  & 0.78 & 0.79 & 0.78 & 0.81 & 0.90 & 0.90 & \textbf{0.90} & 0.90
\\\hline
\multirow{3}{*}{CNN (2,3,4)}& GloVe & 0.81 & 0.79 & \textbf{0.80} & 0.85 & 0.89 & 0.89 & 0.90 & 0.89\\
    \cline{2-10}
    & GloVe \& Syn. \& Emo. & 0.81 & 0.78 & 0.79 & 0.85 & 0.90 & 0.90 & \textbf{0.90} & 0.90\\
    \cline{2-10}
 & GloVe \& Syn. \& Emo. \& Prof. & 0.81 & 0.77 & 0.78 & 0.84 & 0.91 & 0.91 & \textbf{0.91} & 0.91
\\\hline
\multirow{3}{*}{CNN (2,3,4,5)}  & GloVe & 0.83 & 0.78 & \textbf{0.80} & 0.85 & 0.89 & 0.89 & 0.89 & 0.89
\\\cline{2-10}
  & GloVe \& Syn. \& Emo. & 0.80 & 0.77 & 0.78 & 0.84 & 0.89 & 0.89 & \textbf{0.89} & 0.89
\\\cline{2-10}
& GloVe \& Syn. \& Emo. \& Prof.  & 0.81 & 0.79 & \textbf{0.80} & 0.85 & 0.90 & 0.90 & \textbf{0.90} & 0.90
\\\hline
\end{tabular}
\end{small}
\end{table}

Table~\ref{sup_agency_social} shows the 10-fold cross-validation
results for the 10,560 labeled data. The macro-f1 score is reported
along with the Precision, Recall, and Accuracy per label type. The
Logistic Regression model, with profile features, yields F1-score 0.53
for \textit{agency} prediction and 0.56 for \textit{social}
prediction. The CNN with word embedding outperforms  Logistic
Regression with F1-score 0.80 for \textit{agency} prediction and 0.90
for \textit{social} prediction. These results demonstrate that the
happy moment contains enough information for the affective
classification without profile features. The XGBoosted Forest with
syntactic and emotional features also reaches a competitive F1-score of 0.78
for \textit{agency} prediction and 0.90 for \textit{social}
prediction, meaning that these features are as representative as the
word embedding. The different results of \textit{social} and
\textit{agency} imply that the prediction of \textit{social} label
might not rely on the text input, and it's easier to predict. An
additional experiment run to explore this was the addition of the top
1000 unigrams as features for the XGBoosted Forest, however this led
to little (1-2\%) to no increase in predictive power. To further
explore the feature set, we incrementally add the syntactic
features and emotional features, followed by the profile features to the CNN
model. The results show that adding the syntactic and emotional
features leads to a slight drop for \textit{agency}.
Though adding the profile features to the CNN model might
lead to small improvements, we mainly focus on the word embedding,
syntactic features and emotional features for the semi-supervised
learning.

\subsubsection{Semi-Supervised Learning.}
After getting the best models from the supervised learning, we generate the pseudo labels for the 72,324 unlabeled data using the XGBoosted Forest and the CNN models. Then we combine the labeled training data with the pseudo-label data to train the semi-supervised models via 10-fold cross-validation. The validation set is always held out during the training.
Performance of our models are reported in Table~\ref{semi_agency_social}.

\begin{table}
\centering
\caption{10-fold cross-validation semi-supervised learning for \textit{agency} and \textit{social} prediction.}
\label{semi_agency_social}
\begin{small}
\begin{tabular}{| *{12}{c|} }
\hline
\bfseries \multirow{2}{*}{Model} & \bfseries \multirow{2}{*}{Features} & \multicolumn{5}{c|}{\textbf{Agency}}  & \multicolumn{5}{c|}{\textbf{Social}} \\
\cline{3-12}
 & & \textbf{P} & \textbf{R} & \textbf{F1} & \textbf{Acc} & \textbf{AUC} & \textbf{P} & \textbf{R} & \textbf{F1} & \textbf{Acc} & \textbf{AUC}
\\
\hline
XGBoosted Forest & Syn. \& Emo.  & 0.80 & 0.81 & \textbf{0.79} & 0.81 & 0.68 & 0.91 & 0.91 & \textbf{0.91} & 0.91 & 0.91 
\\\hline
\multirow{2}{*}{CNN (2,3,4)}  & GloVe & 0.81 &0.78 &0.79 &0.85 & 0.78 & 0.89 & 0.89 & 0.89 & 0.89 & 0.89
\\\cline{2-12}
 & GloVe \& Syn. \& Emo. & 0.80 & 0.79 & 0.79 & 0.84 & 0.79 & 0.90 & 0.90 & \textbf{0.90} & 0.90 & 0.90 
\\ \hline

\multirow{2}{*}{CNN (2,3,4,5)}  & GloVe & 0.82 & 0.79 & \textbf{0.80} & 0.85 & 0.79 & 0.89 & 0.89 & 0.89 & 0.89 & 0.89
\\\cline{2-12}
 & GloVe \& Syn. \& Emo. & 0.80 & 0.78 & 0.79 & 0.84 & 0.78 & 0.90 & 0.90 & \textbf{0.90} & 0.90 & 0.90 
\\ \hline
\end{tabular}
\end{small}
\end{table}

We had expected performance improvements from semi-supervised
learning, but notice that for the CNN model, the additional 70k
pseudo-labeled data does not improve performance. Compare Table
Table~\ref{sup_agency_social} and Table~\ref{semi_agency_social}.  In
Table~\ref{sup_agency_social} for \textit{agency} prediction, the best
model CNN region (2, 3, 4, 5) with GloVe has an F1-score of 0.80
and in Table~\ref{semi_agency_social} its F1-score remains 0.80. 
Similarly, the best CNN model with embeddings, syntactic and
emotional features for \textit{social} prediction gets an F1-score of
0.90 in Table~\ref{sup_agency_social} as well as
Table~\ref{semi_agency_social}. 

Note also that the XGBoosted Forest provides good performance with
syntactic and emotional features, and its performance improves
slightly after semi-supervised learning, e.g. for \textit{agency}
prediction, it has an F1-score of 0.78 for supervised learning, and
0.79 for semi-supervised learning. For the \textit{social} prediction,
its F1-score is 0.90 for supervised learning and 0.91 for
semi-supervised learning. These results encourage us to investigate
the further impact of syntactic and emotional features
for affective prediction tasks.

\section{Concepts Modeling}

This work extends the modeling procedures described above to
predicting the \textit{concepts} label within the HappyDB. We are
interested in the \textit{concepts} features since they represent the
theme of different types of happy moments. However, we expect that
this is a much more difficult task as it is a multi-class problem. For
the \textit{concepts} modeling task, we are interested in both
improving the prediction results, and interpreting the performance of
the models.

\begin{figure}[h]
\centering
\includegraphics[width=4.2in]{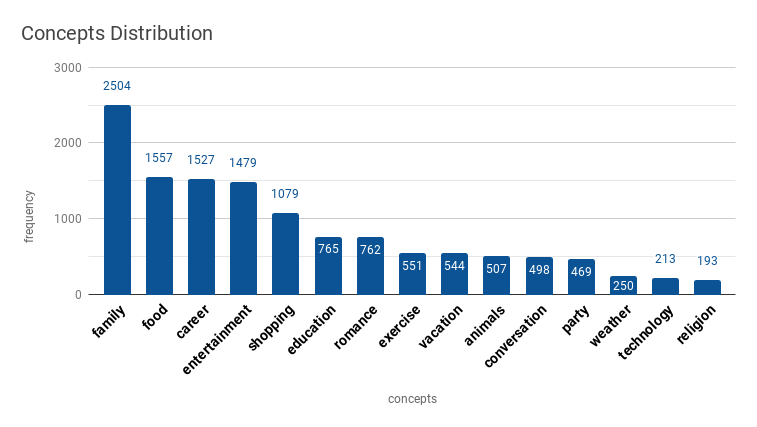}
\caption{The frequency of \textit{concepts} is ordered from highest to lowest from left to right. For example, The most common \textit{concepts} is Family with 2504 examples and the least common \textit{concepts} is Religion with 193 examples. \label{concept-fig}}
\end{figure}

\subsection{XGBoosted Forest Model}

For this task the labeled 10k data set is split into a training set containing 67\% of the data and the remaining 33\% is used for the test set. Within the training set, a 10-fold cross-validation procedure is used.

There are 15 unique \textit{concepts} in the corpus, which are shown in Figure~\ref{concept-fig}, however they are commonly associated with each other as some instances within HappyDB have multiple \textit{concepts} attached. To simplify and examine if \textit{concepts} are distinguishable from each other, we only model the cases where a singular \textit{concepts} tag has been applied. 
Using the same feature set and modeling procedure for the XGBoosted Forest, Table~\ref{concept}a shows the performance of the model on each unique \textit{concepts} tag. The rows of the table are ordered by model performance.

Overall the model shows some promising performance across all \textit{concepts}. The model shows some good performance in the top 3 \textit{concepts}, despite the small sample size for Religion, it appears to be performing the best. However all other \textit{concepts} with lower than 100 instances show much poorer performance. 

One possibility for the poor performance in some of these \textit{concepts} may be the association between them that is already present within the HappyDB, as many \textit{concepts} are used together. Future work could take these common associations and look at potentially making a more hierarchical modeling procedure.

\subsection{CNN Model}

\begin{table}[!htb]
\caption{Concepts Prediction. Concepts are ordered by decreasing F1-score in each case.}
\label{concept}
\begin{subtable}{.5\linewidth}
\caption{(a): XGBoosted Forest}
\label{forest_concept}
\centering
\begin{small}
\begin{tabular}
{@{}|P{2cm}|P{1cm}|P{1cm}|P{1cm}|@{}}
\hline
\bfseries Concept & \bfseries P & \bfseries R & \bfseries F1\\
\hline
Religion & 0.86  & 0.95 & 0.90
\\\hline
Food & 0.81  & 0.86 & 0.83
\\\hline
Family & 0.73  & 0.90 & 0.81
\\\hline
Career & 0.67  & 0.80 & 0.72
\\\hline
Entertainment & 0.70  & 0.75 & 0.72 
\\\hline
Shopping & 0.67  & 0.66 & 0.66 
\\\hline
Animals & 0.59  & 0.58 & 0.58
\\\hline
Romance & 0.65  & 0.48 & 0.56 
\\\hline
Conversation & 0.56  & 0.52 & 0.54
\\\hline
Weather & 0.54  & 0.41 & 0.47
\\\hline
Education & 0.54  & 0.40 & 0.46 
\\\hline
Party & 0.71  & 0.33 & 0.45
\\\hline
Exercise & 0.48  & 0.36 & 0.42
\\\hline
Vacation & 0.50  & 0.34 & 0.40
\\\hline
Technology & 1.00  & 0.11 & 0.20 
\\\hline
\end{tabular}
\end{small}
\end{subtable}
\begin{subtable}{.5\linewidth}
\caption{(b): CNN}
\label{cnn_concept}
\centering
\begin{small}
\begin{tabular}
{@{}|P{2cm}|P{1cm}|P{1cm}|P{1cm}|@{}}
\hline
\bfseries Concept & \bfseries P & \bfseries R & \bfseries F1 \\
\hline
Religion & 0.98  & 0.91 & 0.94
\\\hline
Food & 0.91  & 0.86 & 0.88
\\\hline
Entertainment & 0.88  & 0.86 & 	0.87
\\\hline
Animals & 0.92  & 0.84 & 0.87 
\\\hline
Career & 0.85  & 	0.87 & 0.86
\\\hline
Shopping & 0.88  & 0.84 & 0.86
\\\hline
Family & 0.86  & 0.83 & 0.84
\\\hline
Education &0.88  & 0.73 & 0.79
\\\hline
Weather & 0.88  & 0.72 & 0.77
\\\hline
Exercise & 0.82 & 0.74 & 0.76
\\\hline
Party & 0.89 & 0.70 & 0.76
\\\hline
Vacation & 0.84 & 0.71 & 0.75
\\\hline
Conversation & 0.82  & 0.71 & 	0.75
\\\hline
Romance & 0.77  & 0.64 & 0.68
\\\hline
Technology & 0.83  & 0.59 & 0.64
\\\hline
\end{tabular}
\end{small}
\end{subtable}
\end{table}
Since the CNN model handles multiple classes, we convert the value of \textit{concepts} into a one-hot vector with 15 dimensions, allowing multiple \textit{concepts} attached to a happy moment. We explore the performance of CNN region size (2, 3, 4, 5) with syntactic features and emotional features by 10-fold cross-validation. The overall accuracy of the model is 0.596 and F1-score is 0.629. The metrics for each \textit{concepts} are demonstrated in Table~\ref{concept}b. 

Table~\ref{concept}b shows that the CNN model is generally better than the XGBoosted Forest Model for the \textit{concepts} prediction. For example, the highest F1 for CNN is 0.94 for Religion, and the lowest F1-score is 0.64 for Technology; while the highest F1-score for XGBoosted Forest is 0.9 and lowest is 0.2, meaning that the CNN model is more robust and steady for multi-class prediction. The \textit{concepts} that are improved more than 30\% by CNN are Weather, Party, Education, Exercise, Vacation, and Technology. We suggest that the different performances are caused by the word information that is missing in XGBoosted Forest. The POS and LIWC features used within the XGBoosted Forest appear to be sufficient enough to cover general patterns only within Religion, Food, Entertainment, Career, and Family. 

Besides the above features, we also explored adding the
profile features. Its overall accuracy is 0.601 and F1-score is
0.626.  Adding the profile features to the model doesn't provide
large improvements, but there are small improvements (1\%) for most of
the \textit{concepts}, with the biggest improvement in Exercise which
increased by 3\%. Intuitively, some profile features can be good
markers for \textit{concepts} prediction such as \emph{age},
\emph{married}, and \emph{parenthood}. For instance, young people tend
to discuss events of Education, while the parents are likely to be
happy for a Family theme. On the other hand, the \textit{concepts}
that drop 1\% after adding the profile features are Shopping, Weather,
Party, Conversation. The biggest changes include Technology, whose
F1-score drops by 4\%.

Though these models have different performance, they all illustrate
the difficulty of predicting certain \textit{concepts} and share a
similar prediction trend. The trend is not affected by the 
size of the training data as in  Figure~\ref{concept-fig}. For example,
both models agree that Religion and Food are much easier to predict
than Romance and Technology, but both Religion and Technology have 
small training sets, suggesting that  perhaps Religion contains many
discriminative patterns that make it easier to predict. The
performance also implies that Religion and Food might have some
distinctive patterns with global agreement to represent the happy
moment while Romance and Technology might vary, meaning that people
might have very different views of what causes happiness
in these \textit{concept} themes.

\subsection{Syntactic Pattern Analysis}
To further interpret the performance of the above models, we apply AutoSlog \cite{Riloff96,Wu17}, a weakly supervised linguistic-pattern
learner, to collect the compositional syntactic patterns for the 10k labeled data. Table~\ref{table:high-patterns} illustrates the most frequent syntactic patterns in the data. For each pattern, we list the top 3 \textit{concepts} with the highest probability (no less than 10\%) given the pattern. In the top 15 list, there are 3 patterns include MY (FAMILY MEMBER). This might explain why \textit{social} prediction is easier than the other tasks. As for the \textit{concepts}, a Family theme usually dominates the pattern of MY (FAMILY MEMBER), which implies that when Romance and Family co-occur, the classifier would tend to predict Family, leading to a low recall for Romance.


\begin{table*}[t!h]
\caption{\label{table:high-patterns} Selected top 15 syntactic patterns using AutoSlog-TS Templates. }
\begin{small}
\ra{1.3}
\begin{tabular}{@{}P{0.65cm}|P{4.7cm}|L{6cm}@{}}
\toprule
\bf Freq & {\bf Pattern and Text Match} & \textbf{Concepts Probability and Examples}\\ \hline
1395 & $<$subj$>$ ActVp (\texttt{WENT}) & Family 0.15, Shopping 0.14, Food 0.12
\newline 
\textit{Example: I \textbf{went} for a walk with my wife.}
\\ \hline
1303 & $<$subj$>$ ActVp (\texttt{GOT}) &  Career 0.25, Family 0.17, Food 0.10 
\newline 
\textit{Example: I finally \textbf{got} a job interview.}
\\ \hline
1153 & $<$subj$>$ ActVp (\texttt{MADE}) &  Family 0.23, Food 0.23, Career 0.13
\newline 
\textit{Example: I \textbf{made} a delicious meal.}
\\ \hline
605 & Subj AuxVp $<$dobj$>$ (\texttt{HAVE I}) &   Food 0.32, Career 0.12, Family 0.12 
\newline 
\textit{Example: \textbf{I had} excellent dinner.}
\\ \hline
571 & $<$subj$>$ AuxVp Adjp (\texttt{BE HAPPY}) & Family 0.23, Career 0.13
\newline 
\textit{Example: I \textbf{was happy} to see some friends while they were on vacation.}
\\ \hline
518 & Adj Noun (\texttt{MY HUSBAND}) & Family 0.44, Romance 0.27, Food 0.13 
\newline 
\textit{Example: \textbf{My husband} surprised me with my favorite treats.}
\\ \hline
500 & $<$subj$>$ AuxVp Adjp (\texttt{BE ABLE}) &  Career 0.16, Family 0.13, Shopping 0.12
\newline 
\textit{Example: I \textbf{was able} to get off of work early.}
\\ \hline
495 & Adj Noun (\texttt{MY WIFE})  & Family 0.40, Romance 0.28, Food 0.13
\newline 
\textit{Example: I got a kiss from \textbf{my wife}!}
\\ \hline
476 & ActVp $<$dobj$>$ (\texttt{BOUGHT}) & Shopping 0.61 
\newline 
\textit{Example: I \textbf{bought} a new laptop.}
\\ \hline
458 & Adj Noun (\texttt{MY FAMILY}) &  Family 0.51, Food 0.13, Vocation 0.12
\newline 
\textit{Example: I went on vacation with \textbf{my family}}
\\
\bottomrule
 
 \end{tabular}
\end{small}
\vspace{-.1in}
\end{table*}

Moreover, the \textit{concepts} of Family, Career, Food, and Shopping
contain distinctive syntactic patterns, consistent with
classification performance. To validate our assumptions about the
discriminative level of Technology and Religion, we look at the most
frequent patterns. For example, the most common pattern for Technology
is (BOUGHT) with 36 examples, and Religion's common pattern is (WENT
TO) with 148 examples. Similar to lexical-diversity, we define the
pattern-diversity as the number of unique patterns divided by the
total number of patterns. Technology contains 1078 syntactic patterns,
the average frequency for each pattern is 2, the standard deviation of
frequency is 3, and the pattern-diversity is 0.55, whereas Religion
includes 304 patterns, the average count is 3, the standard deviation
is 10, and the pattern-diversity is 0.51. As for Family, which has
general syntactic patterns, the average count is 3, the standard
deviation is 10, and the pattern-diversity is 0.37. A larger standard
deviation implies more typical patterns, and a smaller
pattern-diversity implies the \textit{concepts} tends to include
stronger syntactic patterns. Since Technology has a similar number of
examples as Religion, this suggests that Religion is more easily
identified because it has many typical syntactic patterns. Our
observation suggests that the syntactic patterns can be strong markers
for affective classification tasks even when the \emph{profile
  features} are missing.


\section{Discussion and Future Work}
We explored the features and models with supervised learning and semi-supervised learning for \textit{social}, \textit{agency} and \textit{concepts} in order to answer some of the questions during the experiments:
\begin{itemize}
    \item \textit{Whether syntactic features and emotional features, which are generated from the text, are representative?} Our experiments show that the syntactic features and emotional features are informative features as they are competitive to word embeddings, and could outperform the word embeddings for some tasks.  

    \item \textit{Whether the profile features are representative? How to convert them into a meaningful granularity?} The profile features include large background information of the writer, therefore, they might represent some general happiness group. 
    However, some features should be mapped to a meaningful group, such as country and age. 
    Though our experiment results are relatively low for the profile features, it could slightly improve the predictions of \textit{social} and some \textit{concepts} when combined with other features. On the other hand, the text input and its extracted features perform well on the prediction task, which 
indicates that the demographic information is not necessary to achieve decent performance. 


    \item \textit{Whether the neural network model outperforms the traditional machine learning model? Whether the best models for \textit{agency} and \textit{social} prediction give good performance for \textit{concepts} prediction?} The CNN models provide promising performance for multi-class classification and semi-supervised learning, while the traditional machine learning method XGBoosted Forest also generates competitive or even better results for binary class prediction and semi-supervised learning. Our best feature sets and models for \textit{social} and \textit{agency} prediction also provide good performance for \textit{concepts} prediction. Our syntactic pattern analysis also demonstrates that these tasks share common characteristics. Therefore, we believe that they are generally robust models for affective classification tasks.
    
\end{itemize}

During the experiments, we realize that the variation and the definition of happy moments might affect the performance of \textit{concepts} modeling, and imply the generalization level of different \textit{concepts}. Some generic characteristics, which are implied by the syntactic and emotional features, are shared between the classes of \textit{agency}, \textit{social} and \textit{concepts}. The linguistic-pattern learner AutoSlog provides insightful syntactic patterns for us to interpret the models. In future work, we will focus on utilizing syntactic patterns for other affective classification tasks and identify common patterns. We also hope to explain such patterns and the \textit{concepts} theme with psychological theories. Finally, we are curious to see if there are generic characteristics or common compositional semantic patterns for modeling happy or unhappy moments with cross-domain data.

%
%
%
%

\end{document}